\documentclass[letterpaper, 10 pt, conference]{ieeeconf}  

\IEEEoverridecommandlockouts                              

\title{\LARGE \bf
FingerTac - An Interchangeable and Wearable Tactile Sensor \\for the Fingertips of Human and Robot Hands
}

\author{Prathamesh Sathe$^{1}$ Alexander Schmitz$^{1}$  Satoshi Funabashi$^{1}$ \\Tito Pradhono Tomo$^{1}$ Sophon Somlor$^{1}$ and Sugano Shigeki$^{1}$ 
\thanks{This work was not supported by any organization.}
\thanks{$^{1}$Prathamesh Sathe, Alexander Schmitz, Satoshi Funabashi, Tito Pradhono Tomo, Sophon Somlor and Shigeki Sugano are with Waseda University, Tokyo, Japan (corresponding author e-mail: sathe.prathamesh@fuji.waseda.jp)}
}

\usepackage{amsmath}

\usepackage{epsfig}
\usepackage{epstopdf}
\usepackage{subcaption}
\usepackage{dblfloatfix}
\usepackage{cite}
\usepackage{soul}
\usepackage{xcolor}
\usepackage{graphicx}
\usepackage{printlen}
\usepackage{algpseudocode}
\usepackage{algorithm}
\usepackage[bookmarks=false]{hyperref}

\begin{document}

\maketitle
\thispagestyle{empty}
\pagestyle{empty}

\begin{abstract}
Skill transfer from humans to robots is challenging. Presently, many researchers focus on capturing only position or joint angle data from humans to teach the robots. Even though this approach has yielded impressive results for grasping applications, reconstructing motion for object handling or fine manipulation from a human hand to a robot hand has been sparsely explored. Humans use tactile feedback to adjust their motion to various objects, but capturing and reproducing the applied forces is an open research question.  
In this paper we introduce a wearable fingertip tactile sensor, which captures the distributed 3-axis force vectors on the fingertip.
The fingertip tactile sensor is interchangeable between the human hand and the robot hand, meaning that it can also be assembled to fit on a robot hand such as the Allegro hand. This paper presents the structural aspects of the sensor as well as the methodology and approach used to design, manufacture, and calibrate the sensor. 
The sensor is able to measure forces accurately with a mean absolute error of 0.21, 0.16, and 0.44 Newtons in X, Y, and Z directions, respectively.

\end{abstract}

\section{Introduction}

Researchers use different ways to teach robots new skills. In direct teaching \cite{Direct_Teaching_Example-1} the user moves the robot hand within the robot's kinematic framework to demonstrate a task. This is the easiest way for robots to learn new skills such as pick and place tasks. 
However, direct teaching cannot be employed for teaching the robots to perform multi-fingered manipulation tasks, as it is not possible for the user to move the robot fingers simultaneously through manual inputs to handle objects.

Teleoperation is another possibility. The user teleoperates the robot hand with a data glove which measures the joint angle data of the user's finger joints. Teleoperation has empowered researchers to address human-like robot in-hand manipulation and object handling tasks\cite{Teleoperation_Example-1}. During teleoperation, providing force feedback to the user while manipulating a certain object is essential for many tasks. However, providing haptic feedback is challenging, limiting the scope of teleoperation for skill transfer. 

A third approach to teaching robots is through observing human motion. In the case of \cite{dillmann_teaching_2004} and \cite{kang_toward_1995}, the researchers attempt to teach the robot to manipulate different objects by capturing hand movements from a human. This technique of teaching a robot is commonly called ''skill transfer''.

A common limitation in all the above mentioned methods of robot teaching is that only angular position data from the human hand is provided as an input to the robot during training, whereas, force data is not provided. Since humans handle objects of various shapes, sizes, and with different physical properties such as soft and fragile objects, object handling also relies on distributed force sensing. Not only the amount of force, but also the direction of force (including shear forces) is necessary for working in an unstructured environment.

Many of the presently available wearable tactile sensors can only measure normal force but measurement of shear force is not possible.
Furthermore, grasping and manipulating objects is not limited to the palmar side of the fingertips. 
However, most currently available wearable fingertip tactile sensors, especially those that can measure shear force, function as a single taxel (tactile sensing element).
Therefore, they cannot measure the distributed forces applied at different points around the surface of the fingertip. Furthermore, the mechanical design of most of the tactile sensors reduces the ability to measure forces on the lateral side of the fingertip. 

To address the functional and mechanical limitations of previously designed tactile sensors, in this paper we present a wearable fingertip tactile sensor, named FingerTac. It is based on 20 distributed 3-axis Hall-effect sensors around the palmar and lateral side of the fingertip. FingerTac has an anthropomorphic shape 
and allows the user to perform complex tasks such as rolling an object over a wide surface area of the fingertip. 

The rest of this paper is organized as follows. Section II reviews related work on wearable sensors. Section III describes the design of FingerTac, i.e. the ergonomic positioning of the taxels around the fingertip, the mechanical design of the sensor skin along with the flexible PCB design. Lastly this section also highlights the manufacturing process used to manufacture the tactile sensor. Section IV explains the experimental setup and the procedure for calibrating the tactile sensor to measure 3-axis force while applying forces on the fingertip surface and evaluation of the force measured by the tactile sensor. Therefore, we will compare the results of the measured force with a reference force-torque sensor. Section V presents the calibration results. Section VI draws conclusions and discusses possible future work.

\section{Related Works}
The concept of a wearable tactile force sensor was published in \cite{beebe_silicon-based_1998}. The wearable device consisted of a single piezo-resistive tactile sensor wrapped around the fingertip. To safely distribute the force applied on the sensor while handling an object, a dome was designed which covered the top surface of the sensor. Even though the concept of a wearable tactile sensor was innovative, the sensor comprised of only a single taxel which covered most of the contact area on the palmar side of the fingertip. One of the limitations of the sensor was that it could only measure normal force. Also the mechanical form factor of the sensor prevented the sensor from being distributed over different areas around the fingertip. This limited the potential of the sensor to be used in measuring tactile forces generated while performing grasping tasks only.

A wearable tactile sensor was introduced by \cite{Huang_robio2017}. The sensor consists of a piezoelectric material called PDMS and is worn on the user's palmar side of the fingertip. When the PDMS structure is mechanically deformed due to external force, electric charges are generated within the material. The generated values of the charges were calibrated to measure force by using a neural network. Although the authors claim to have good results at measuring forces accurately, it must be noted that the sensor is bulky in construction and as a whole functions as a single taxel. Also, due to the large form factor of the sensor, it cannot be distributed over the entire surface of the fingertip. This again limits the potential of the sensor to be used while performing grasping tasks only. 

The concept of a fingernail tactile sensor based on the principle of photo-plethysmography was introduced in \cite{mascaro_1999}\cite{mascaro_measurement_2004}. The sensor was improved further in \cite{grieve_calibration_2014}\cite{Grieve_2015}. The sensor system consisted of several photodiodes and LEDs. When force was applied on the object, the change in the coloration of the fingernail was observed. The change in the coloration was calibrated to measure force in 3D space. It should be noted that as the sensor characteristics depend highly on the behavior of the biological matter, in this case blood within the fingertips, the performance of the sensor may be altered for different users. This reduces the scope for this sensor to be used for applications like skill transfer. Furthermore, only one force vector per fingertip could be obtained.

\cite{Battaglia2015}\cite{Battaglia2018} presented ThimbleSense, a wearable tactile force sensor which is manufactured by embedding industrial grade force-torque sensors. In both devices, the force-torque sensors were encapsulated within a cap-like structure which is meant to be worn on the finger while manipulating an object. 
The paper suggests that the purpose of developing such a device was to measure accurate tactile force measurement and study sensorimotor control during grasps and manipulation while applying forces and manipulating an object. 
Though this method could accurately measure forces while manipulating different objects, the mechanical design of this sensor is not anthropomorphic in nature. Only a single force-torque sensor is used, and therefore only one force vector per fingertip could be obtained, albeit including also torques.

A wearable multi-modal sensing glove was presented in \cite{Bianchi2016}. The sensors embedded in the glove measure the pressure acting on the different parts of the hand. The tactile sensor used is a piezo resistive fabric sensor which is spread across various parts of the finger to measure tactile forces. Along with the tactile information the wearable device also measures hand pose while performing grasping tasks on various objects. An interesting use case of liquid metal to be used as a tactile sensor was presented by \cite{Wang_2020}. Along with tactile information the sensor is also able to measure temperature. The sensor is made by sandwiching Galinstan liquid between two layers of PDMS material and can act as a tactile sensor to measure force in the normal direction. The material can withstand high degrees of bending and stretching. However it should be noted that the form factor of the sensor limits this sensor to be distributed over the entire surface of the fingertip. Also, as mentioned previously, human object handling relies not only on normal force but also on shear force, thus this limits the use of these devices to quantify a holistic view of object handling and grasping. 

A wearable tactile glove is presented by \cite{Nature_2019}. The glove features 584 closely knitted sensors distributed across the entire hand. The sensors use piezo-resistance as their working principle and are used to measure normal force only. The authors claim to have higher accuracy for heavier objects. Accuracy of measuring force lowers as the weight of the object decreases. This implies that the glove might not be a favorable option to use for measuring forces which are present in fine manipulation tasks.

In the past our lab developed a series of unobtrusive tactile sensors for measuring shear and normal forces applied on the human fingertips while handling objects \cite{harris_iros_2018}\cite{harris_robio_2019}\cite{harris_humanoids_2020}\cite{prathamesh_iros_2020}. Although, the latest iteration of the sensor \cite{prathamesh_iros_2020} provided good results and was effective at measuring normal and shear forces, the sensor functioned as a single taxel. Thus the sensor was incapable at acquiring distributed force over the entire surface of the fingertip. Also, the lever design which is illustrated in Fig 1. of \cite{prathamesh_iros_2020} mechanically limits the user from performing tasks with the sides of the fingertips.

Previously in our lab we have successfully designed and manufactured a Hall-effect based tactile sensor \cite{Tito_2015}. We have developed this sensor further and implemented it to not only cover the robot fingers phalanges \cite{Tito_2016} but also robot fingertips \cite{tito_RAL_2016}. To combat the cross talk and to increase sensor response we have also experimented by changing the skin structure of the sensor \cite{tito_RAL_2018}. Tactile skin sensors based on the Hall-effect principle are an effective and proven way to measure 3-axis force. The bump design as described in \cite{tito_RAL_2018} if combined with a flexible PCB can be modeled into various shapes or forms and thus in this case can be distributed over the entire surface of the fingertip of the human hand as well as the robot fingertip. This would increase the scope of the sensor to be used to measure tactile data in seemingly simple yet complex tasks such as rolling an object over the surface of the fingertip.

\section{Sensor Design}
\begin{figure}[thpb]
    \centering
    \includegraphics[width=\columnwidth]{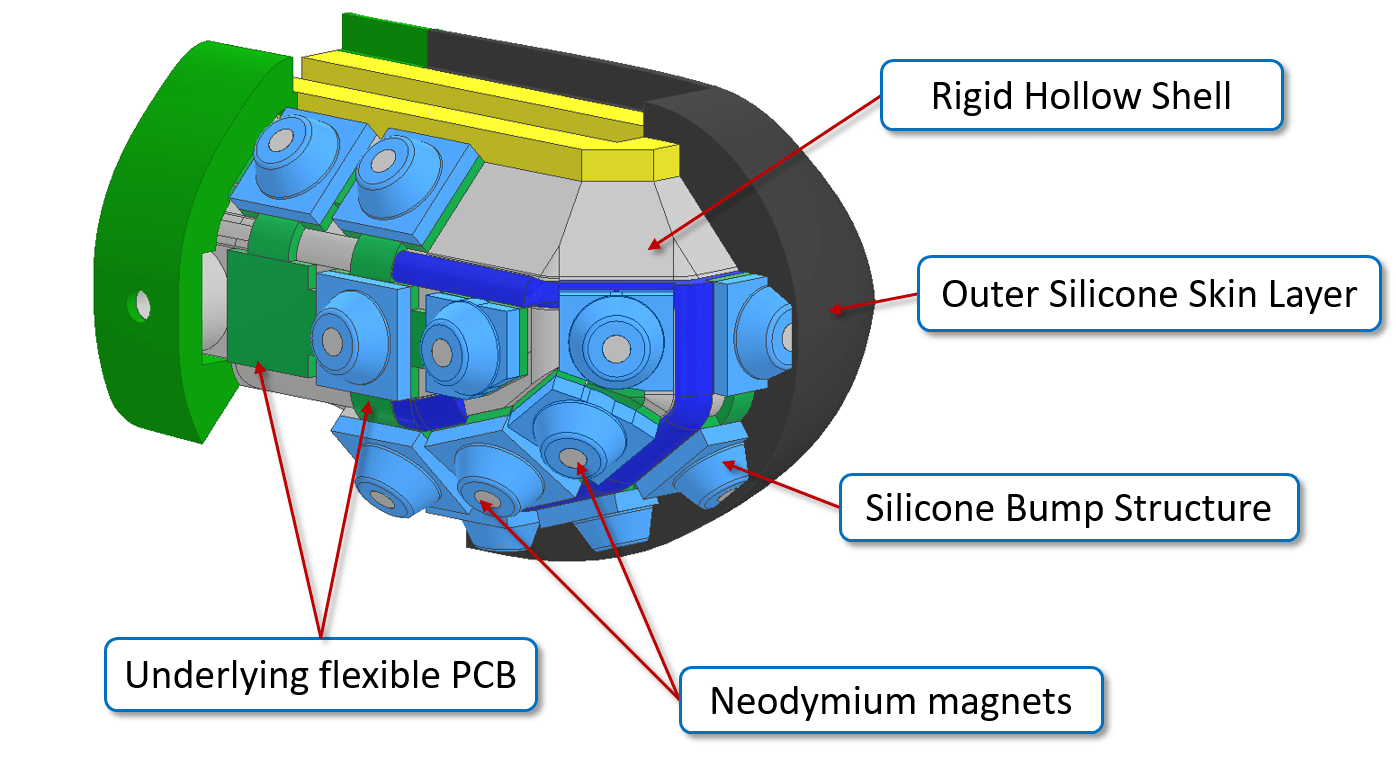}
    \caption{FingerTac: Fingertip Tactile Sensor}
    \label{FingerTac}
\end{figure}

The FingerTac comprises of a composite silicone skin layer which consists of an inner layer of hard silicone bump structures each housing a neodymium magnet and an outer layer of soft silicone skin as shown in Fig.~\ref{FingerTac}. Each bump structure is placed on top of a Hall-effect sensor. 
In total there are 20 Hall-effect sensors mounted on an underlying flexible PCB as shown in Fig.~\ref{FingerTac}.
The PCB is wrapped around a thin but rigid 3D-printed hollow shell designed to be worn on the fingertips as shown in Fig.\ref{FingerTac}. The unique arrangement of bump structures and the overall shape of the FingerTac gives the user the ability to perform not only grasps but also increases the scope of the sensor to be used for measuring tactile data across several points over the surface of the fingertip while performing complex motions such as traversing an object over the surface of the fingertip. 
Before wearing the sensor the user is asked to wear a finger sleeve which acts as a friction surface between the fingertip tactile sensors and the fingertips preventing the sensor from slipping off of the fingertip while performing grasps. The overall size of FingerTac is length: 39~mm, width: 27~mm and height: 26~mm.

\subsection{Sensor Design Criteria}
We followed the following design criteria for FingerTac:
\begin{itemize}
\item Ergonomic localization of taxels around the fingertip.
\item Designing a rigid shell around the fingertip for placing the taxels.
\item Deriving a Flexible PCB from the rigid fingertip shell to accommodate the sensors.
\item Realizing the silicone bump design and composite silicone skin.
\end{itemize}

Each of the above mentioned design steps is explained below in detail:

\subsubsection{Ergonomic localization of taxels around the fingertip}
\begin{figure}[thpb]
    \centering
    \includegraphics[width=\columnwidth]{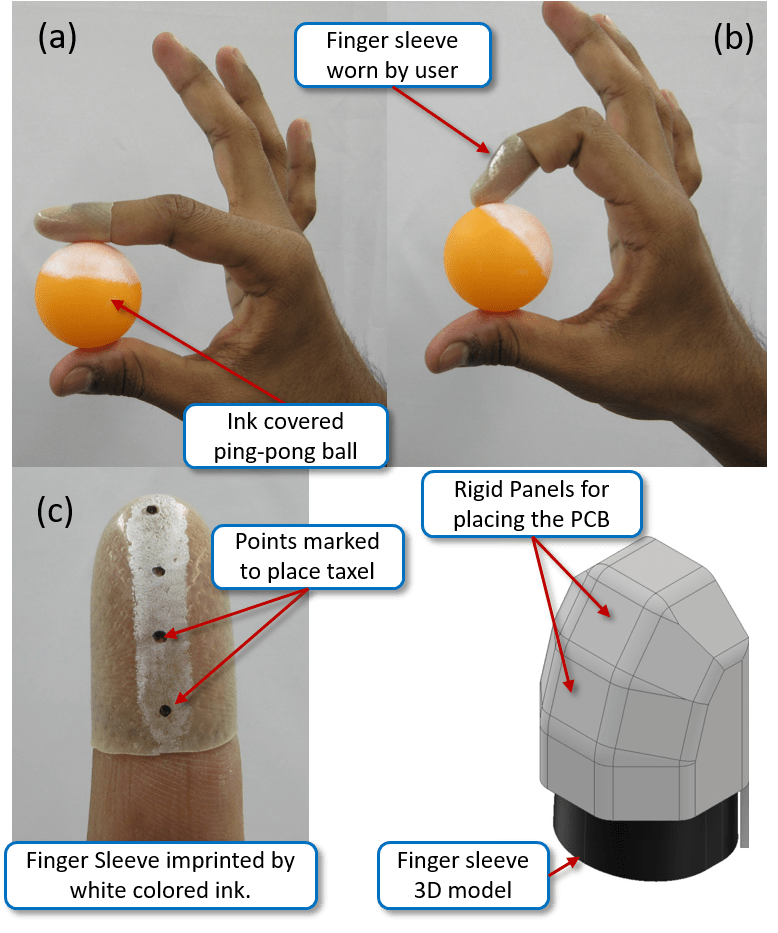}
    \caption{Preliminary object interactive study}
    \label{Study}
\end{figure}

To spatially distribute the taxels over the palmar and lateral sides of the fingertip, a small object interaction study was conducted. The user was told to wear the finger sleeves on top of the fingertips as shown in Fig.~\ref{Study}.
Objects suitable for two point and three point grasp were chosen and the surface of the object was painted with ink. The user was then asked to hold the object in the natural grasping position and perform slight movements while holding the object. For example, in case of holding the ping-pong ball, the user was asked to roll the ping-pong ball over the fingertip area. As the ink covered ping-pong ball traversed the fingertip surface, the ink would be imprinted on the finger sleeve. The imprinted, inked surface on the finger-sleeve indicated the contact area of the fingertip used while moving the ping pong ball over the surface of the fingertip as shown in Fig.~\ref{Study}.

\subsubsection{Designing a rigid shell around the fingertip for placing the taxels}
After observing the contact area indicated by the inked surface on the finger sleeve, dots were marked to indicate the potential positions of the taxels. A CAD model of the fingertip was derived by taking the finger sleeve dimensions. Potential positions of each taxel were marked in the CAD model. 


To house the sensors, all the taxel positions were mapped one by one and rigid sensor panels were designed around the finger sleeve structure. The sensor panels were connected amongst one another and a rigid shell to house all the taxels around the finger sleeve was designed. 

\begin{figure}[thpb]
    \centering
    \includegraphics[width=\columnwidth]{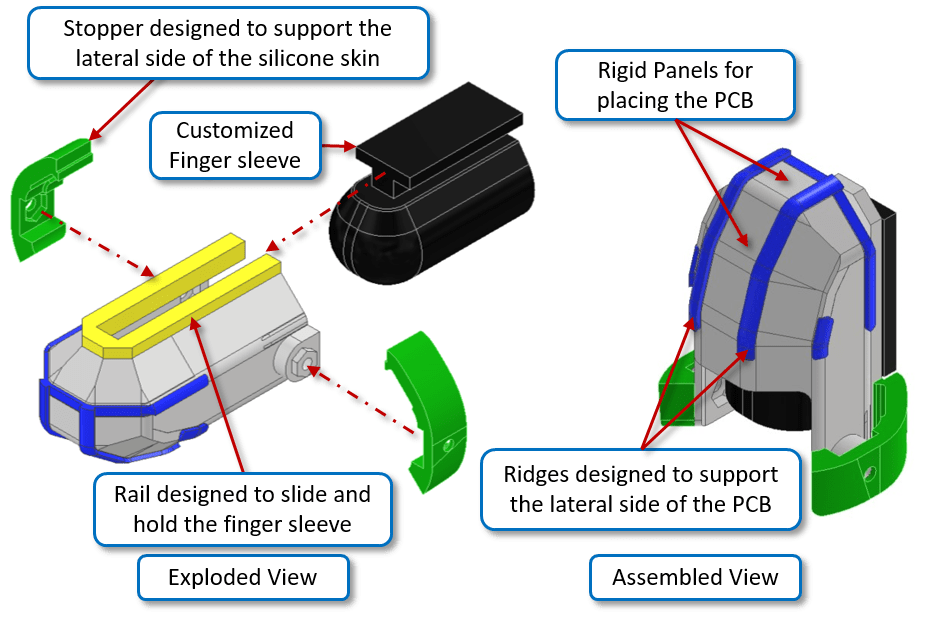}
    \caption{Rigid shell design for placing the sensors}
    \label{RigidShell}
\end{figure}

To restrict the PCB's lateral movement, ridges were designed to hold the PCB in place. The ridges also functioned as a reference for gluing the PCB securely on top of the rigid hollow shell as shown in Fig.~\ref{RigidShell}. To support the lateral side of the silicon, a stopper was designed. The stopper also acted as a frame for the silicone as shown in Fig.~\ref{Mold2} during the molding process. A custom finger sleeve was designed to be worn by the user. The finger sleeve acted as a friction layer between the user's fingertip and the rigid hollow shell. The sleeve was attached to the hollow shell by sliding it inside the rail of the hollow shell, as shown in Fig.~\ref{RigidShell}.

\subsubsection{Deriving a flexible PCB from the rigid fingertip shell to accommodate the sensors}
\begin{figure}[thpb]
      \centering
      \includegraphics[width=\columnwidth]{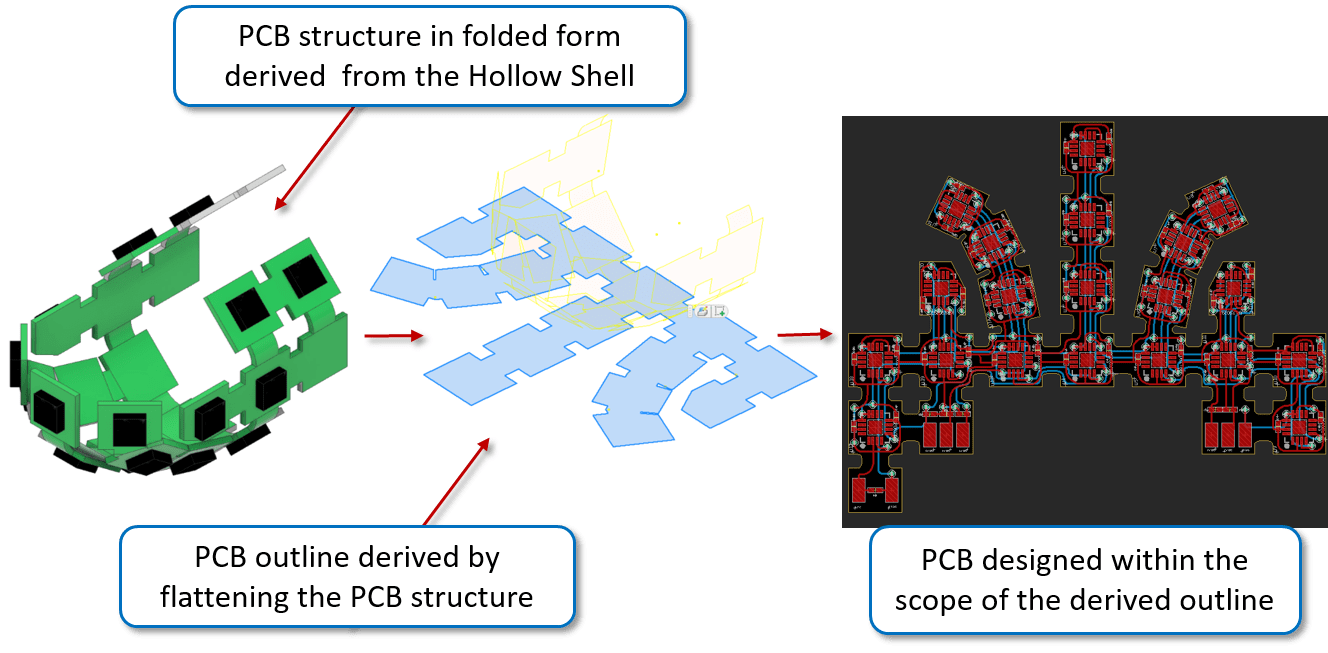}
      \caption{Flexible PCB shape derived from rigid fingertip shell}
      \label{Flexible}
   \end{figure}

A flexible PCB was designed within the confines of the PCB shape derived by flattening the rigid shell as shown in Fig.~\ref{Flexible}.
Once the PCB outline was derived from the flattened PCB, the circuitry confined to the scope of the outline was designed. The flexible PCB comprised of 20 MLX90393 Hall-effect sensor ICs along with essential components such as registers and capacitors.
The I2C communication protocol was used to communicate between the microcontroller and each Hall-effect sensor. The 20 Hall-effect sensors were divided amongst 5 SDA lines such that 4 sensors were connected to one SDA line each.
 The thickness of the flexible PCB was 0.4mm, which allowed it to bend with ease. The PCB could be bent at a maximum angle of 180 degrees. This allowed the PCB to wrap around the hollow shell.
  
\subsubsection{Realizing the silicone bump design and composite silicone skin}
\begin{figure}[thpb]
      \centering
      \includegraphics[width=\columnwidth]{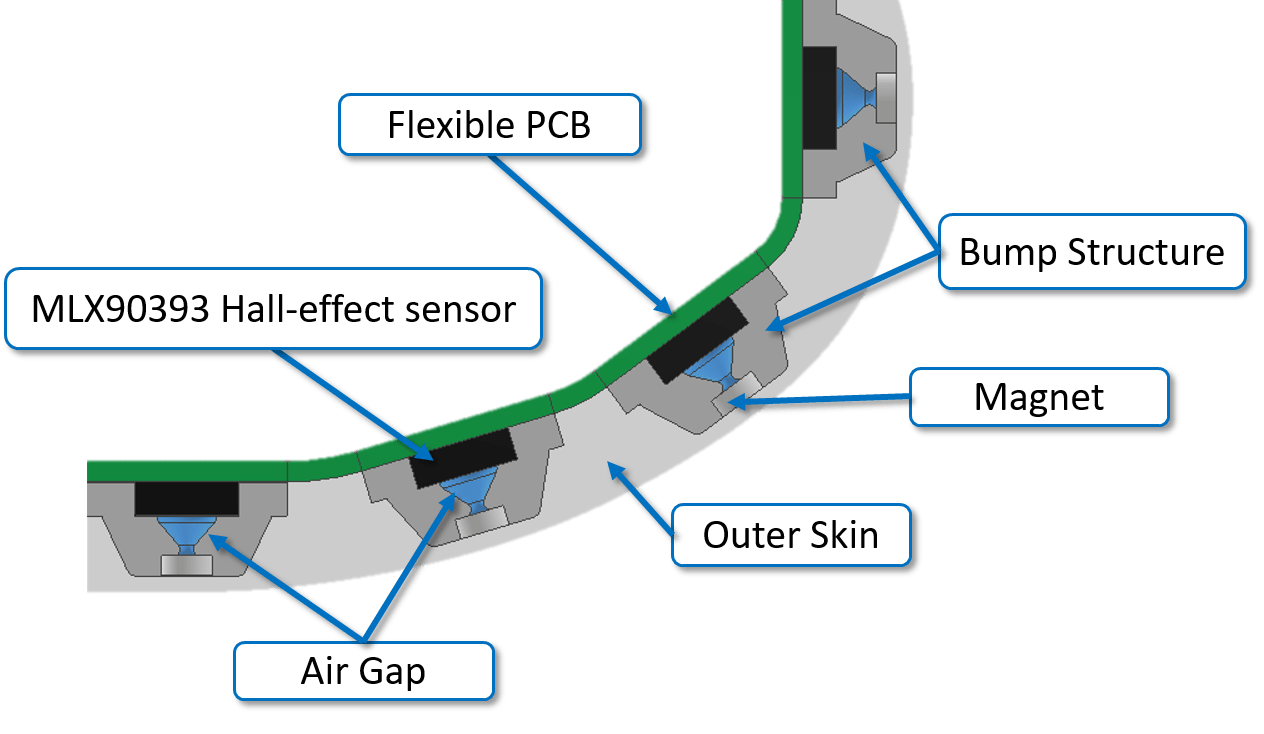}
      \caption{Cross-section of composite silicone skin}
      \label{CompositeSkin}
   \end{figure}

A frustum-shaped silicone bump structure was designed with reference to the previous bump structure design \cite{tito_RAL_2018}. 
The bump structure comprises of an air gap with space on the top of the structure to house the magnet as shown in Fig.~\ref{CompositeSkin}. 
The orientation of the magnet is such that the north side of the magnet faces the Hall-effect sensor. 
The air gap between the magnet and the Hall-effect sensor is 1.2~mm. The overall diameter of the bump structure is 3.1~mm. The base of the structure was molded to the shape of the Hall-effect sensor. 

   
To have a smooth surface for FingerTac, an outer continuous layer was molded over the bump structure.
The continuous skin layer has a thickness of 0.3~mm from the face of the magnets and fills the gap created between two bumps. The layer protects the magnets from external abrasions. Together the soft outer silicone layer as well as the silicone bumps function as a composite skin to distribute the incoming force imparted by the object during object handling to the magnets. 

The outer skin layer was molded with the Ecoflex 50 silicone rubber from Smooth-On, which is softer than the material used for the silicone bumps. Ecoflex 50 has a shore hardness of 00-50. The Silicone bumps were molded with a material called Sil-Poxy from Smooth-On. SilPoxy is a harder silicone rubber with a shore hardness of 40A. 

\subsection{Manufacturing Process}
\subsubsection{Molding of the silicone bump structure}

Once the silicone structure design was finalised, 
a mold comprised of two hollow blocks was designed. A rubber tube of diameter 5~mm was used to insert the liquid silicone. A vacuum assisted molding process was followed to mold the silicone skin. The vacuum assisted molding prevents the formation of bubbles within the silicone during the molding process.
  
  \begin{figure}[thpb]
      \centering
      \includegraphics[width=\columnwidth]{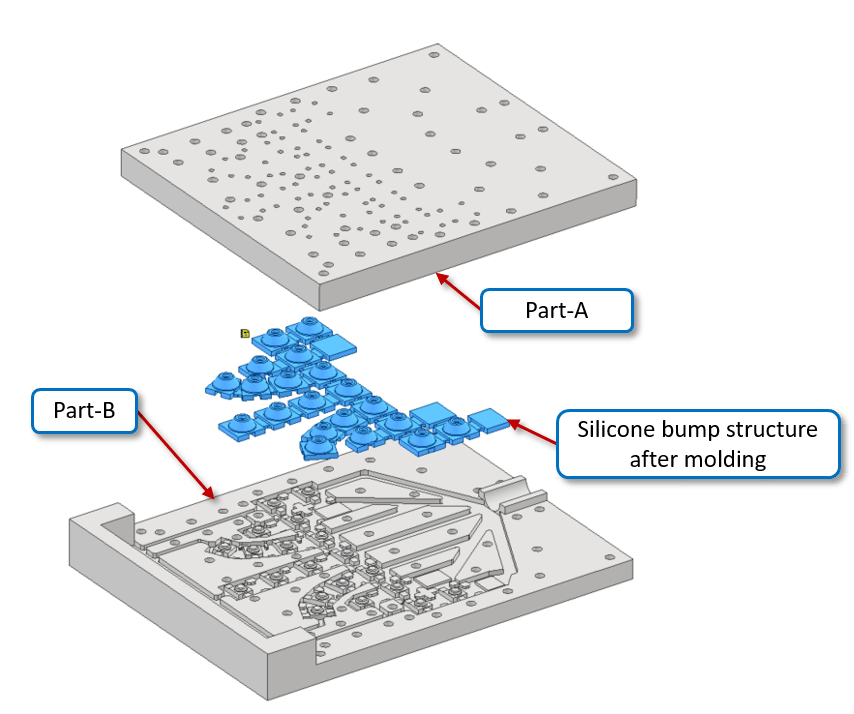}
      \caption{Silicone bump structure mold}
      \label{Mold1}
   \end{figure}

\subsubsection{Molding of composite silicone skin}
\begin{figure}[thpb]
      \centering
      \includegraphics[width=0.9\columnwidth]{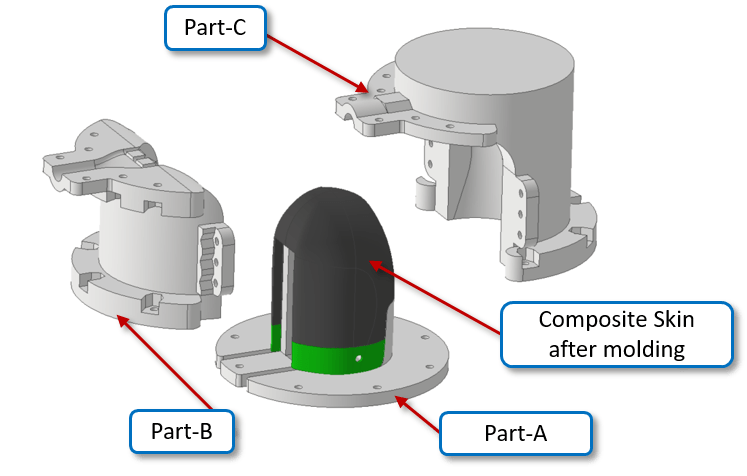}
      \caption{Fingertip sensor mold}
      \label{Mold2}
   \end{figure}

Similar to the mold designed for the silicone structure, three blocks, Part-A, Part-B, and Part-C in the shape of the outer skin layer were designed 
as shown in Fig.~\ref{Mold2}. 
A black color pigment was added to the Ecoflex50, as well as SLIDE from Smooth-On, a liquid surface tension diffuser. It is an additive that creates a cured silicone with greatly reduced surface tension. 
   
\subsection{Wearability of FingerTac}
\begin{figure}[thpb]
      \centering
      \includegraphics[width=0.95\columnwidth]{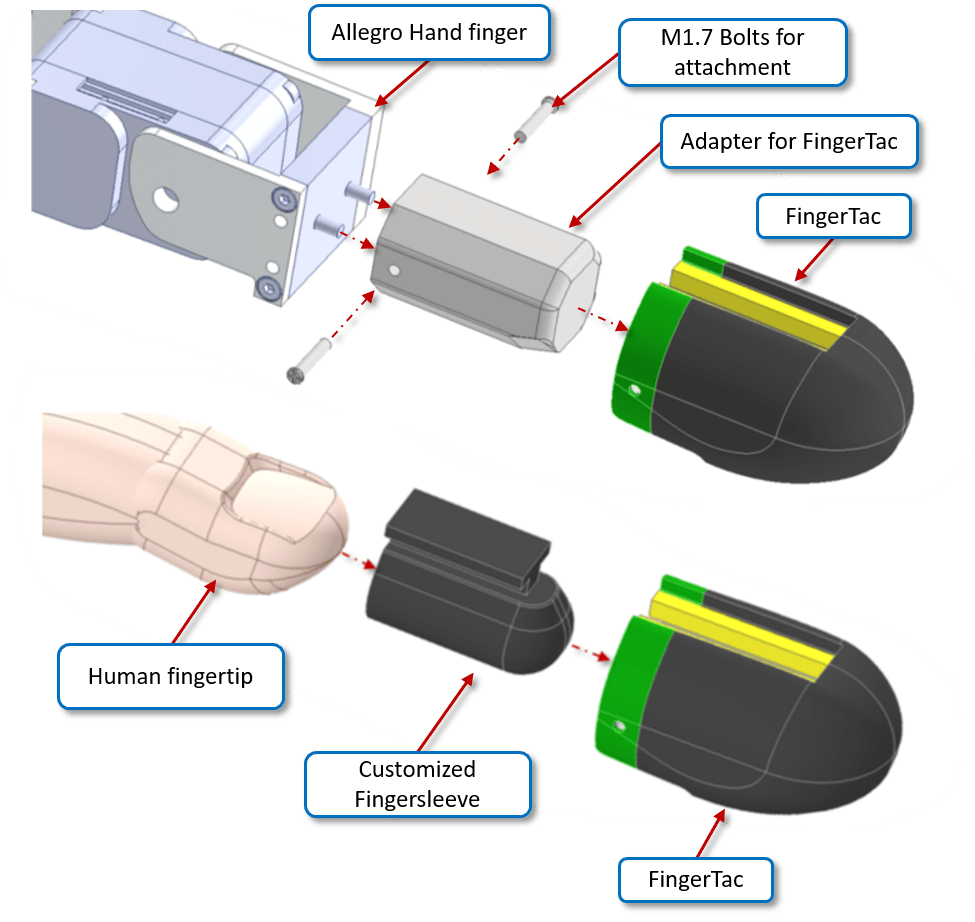}
      \caption{Wearability of FingerTac}
      \label{Wearability}
   \end{figure}

FingerTac can be worn on a human fingertip as well as the fingertip of the Allegro hand as shown in Fig. \ref{Wearability}. 
In the case of attaching the FingerTac on the Allegro Hand, a simple adapter was designed to fit inside the hollow shell of the FingerTac. FingerTac is fastened to the adapter with the help of M1.7 bolts. 

\subsection{Usage of FingerTac while manipulating an object}

\begin{figure}[thpb]
      \centering
      \includegraphics[width=\columnwidth]{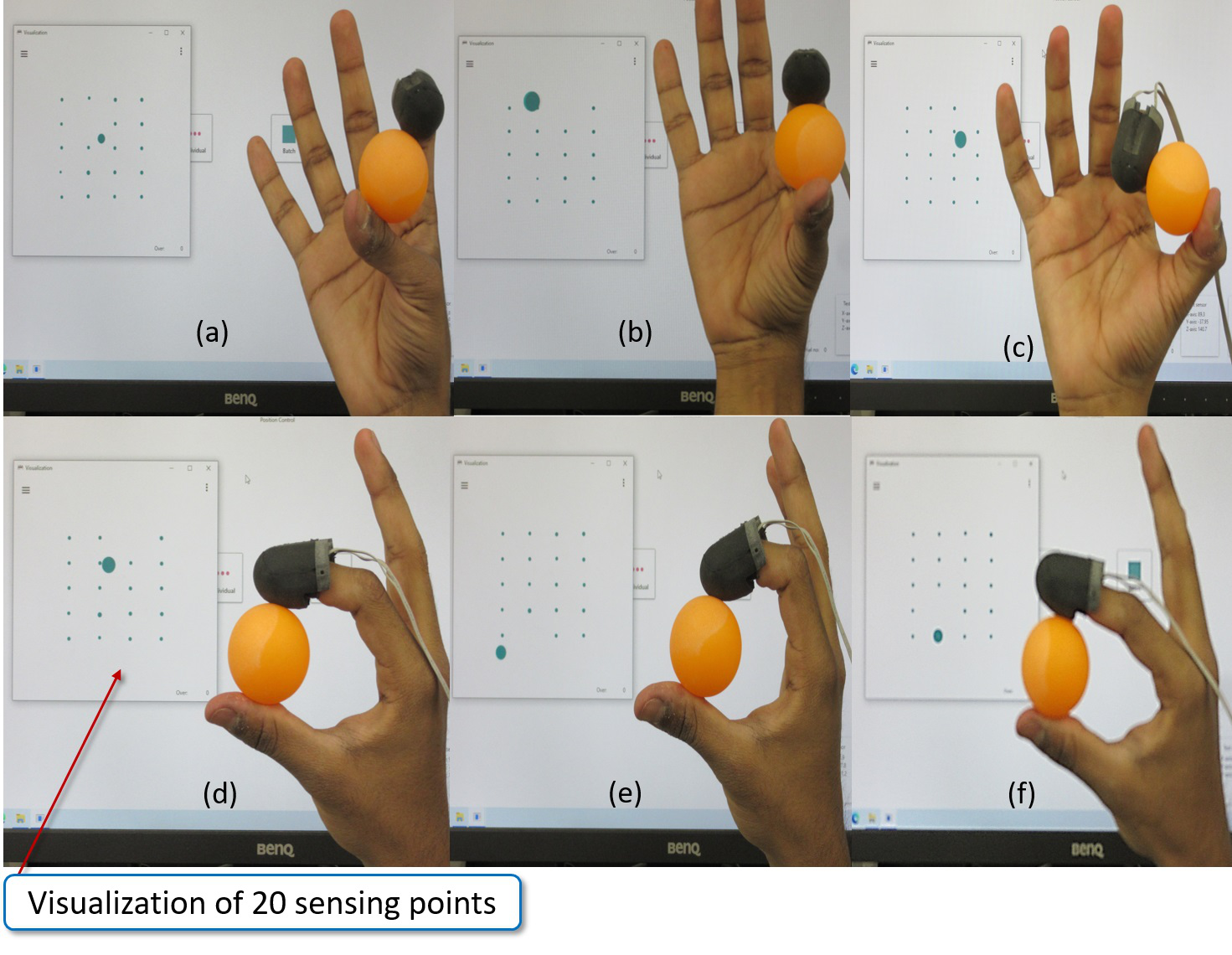}
      \caption{FingerTac measuring tactile data around the surface of the fingertip. In each orientation, a sensing point is active. The arrangement of the sensing points in the visualization is arbitrary. From a-c rolling motion is performed. From d-f pitching motion is performed.}
      \label{Around}
   \end{figure}

FingerTac can be worn on the fingertip of the user to measure tactile data around the surface of the fingertip, see Fig. \ref{Around}. In the figure we can see that the user performs a two fingered manipulation using a ping-pong ball.
\section{Sensor Calibration}
Unlike the previous design of the tactile sensor \cite{prathamesh_iros_2020}
, the current fingertip tactile sensor’s sensing principle does not depend upon physical or biological characteristics such as the hardness or softness of the human fingertip. 

This is because the displacement of the magnet is directly dependent upon the application of an external force. The user’s fingertip does not come in direct contact with the object. As there are no human biological characteristics at play to measure the force applied on the fingertip, the sensor calibration need not be performed each time the user wears the fingertip.

Instead of computationally modeling the mechanical behavior of the bump structure under the application of the force, we have used supervised learning to calibrate the sensor to 3-axis force values by converting the Hall-effect sensor raw data generated due to the application of the external force. 

To calibrate the fingertip tactile sensor, a reference sensor is necessary to train and test the tactile sensor. Two data sets were taken for performing calibration; first, training data for curve-fitting and generating a calibration matrix for predicting force values. Second, test data which acts as an input to estimate force values by using the previously generated calibration parameters. The fingertip tactile sensor is evaluated by comparing the estimated force values to the true force values measured by the reference sensor. We used linear regression with a quadratic model quadratic linear regression model to convert the raw readings from the Hall-effect sensor to 3-axis force measurements. We worked with linear models during preliminary training. However, the estimated force values on the test data were not at par with the true force values measured by the reference sensor.

\subsection{Calibration Setup}
\begin{figure}[t]
      \centering
      \includegraphics[width=\columnwidth]{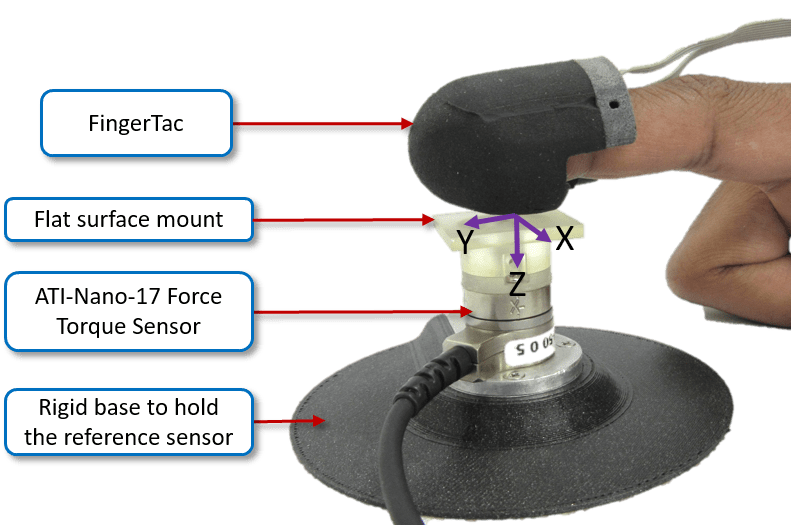}
      \caption{Calibration Setup}
      \label{Setup}
   \end{figure}

The setup for calibration and testing is shown in Fig.~\ref{Setup}. An industrial grade 6-axis force-torque sensor (ATI-Nano17 from BL Autotech) was used as a reference sensor to measure the normal and the shear force indirectly imparted by the user's fingertip. In this paper, only 3 axis forces were used for the calibration of the fingertip tactile sensor. A 3D-printed base plate as shown in the figure was used to hold the force-torque sensor. The base plate was mechanically fastened on a rigid table such that there is no relative motion between the table and the force-torque sensor. As shown in the figure, a 3D-printed flat surface was attached to the top of the 6-axis force/torque sensor. This flat surface provides enough area for the user to comfortably apply normal and shear forces while wearing the fingertip tactile sensor. 

\begin{figure}[thpb]
    \centering      
    \includegraphics[width=0.8\columnwidth]{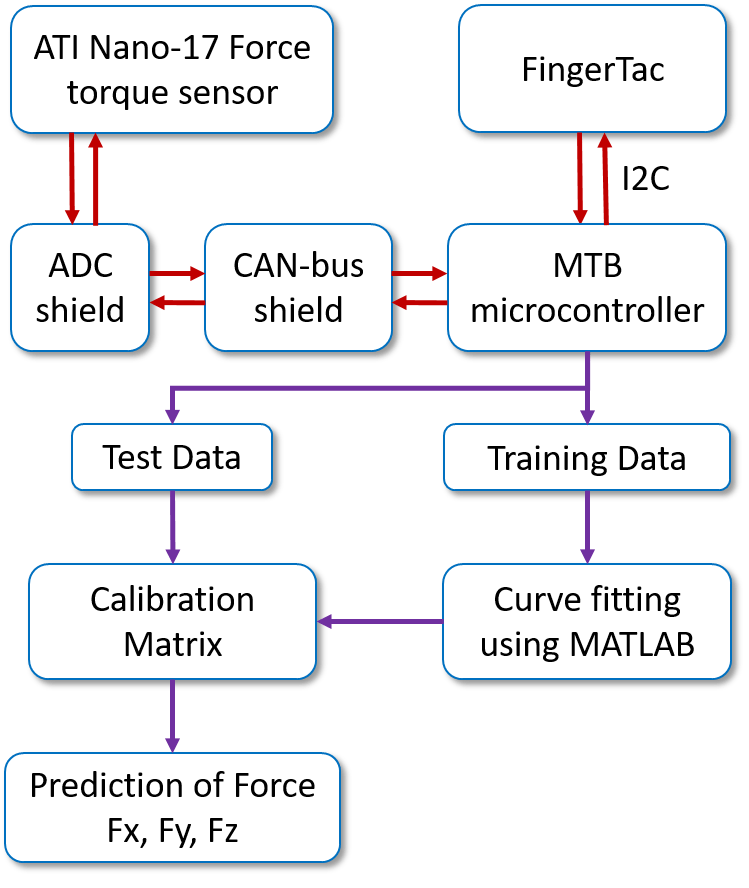}
    \caption{Overview of 3-axis Force calculation}
    \label{ForceCalculation}
\end{figure}

Fig.~\ref{ForceCalculation} shows the overview of the calibration system in form of a simple block diagram. The Hall-effect sensors embedded within the flexible PCB of the fingertip tactile sensor send raw digital tactile data to a microcontroller via the I2C communication protocol. The microcontroller used was an MTB4 by IIT (Italian Institute of Technology). 
The reference sensor’s analog data is first converted to digital data by using an ADC. The digital output data from the ADC is sent to the microcontroller via the CAN-bus communication protocol. By using the CAN-bus the incoming data from the MLX90393 and the ATI Nano-17 can be synced. 
Further, the digital data for each individual axes from the fingertip tactile sensor and the reference force-torque sensor’s force measurements are sent from the microcontroller to the PC via a CAN-bus to Serial interface.
The readings from the MLX90393 and the ATI Nano-17 are collected at a sampling rate of 100Hz. The readings are plotted against time for Calibration. 

\subsection {Calibration Methodology}
\begin{figure}[h]
      \centering
      \includegraphics[scale=0.4]{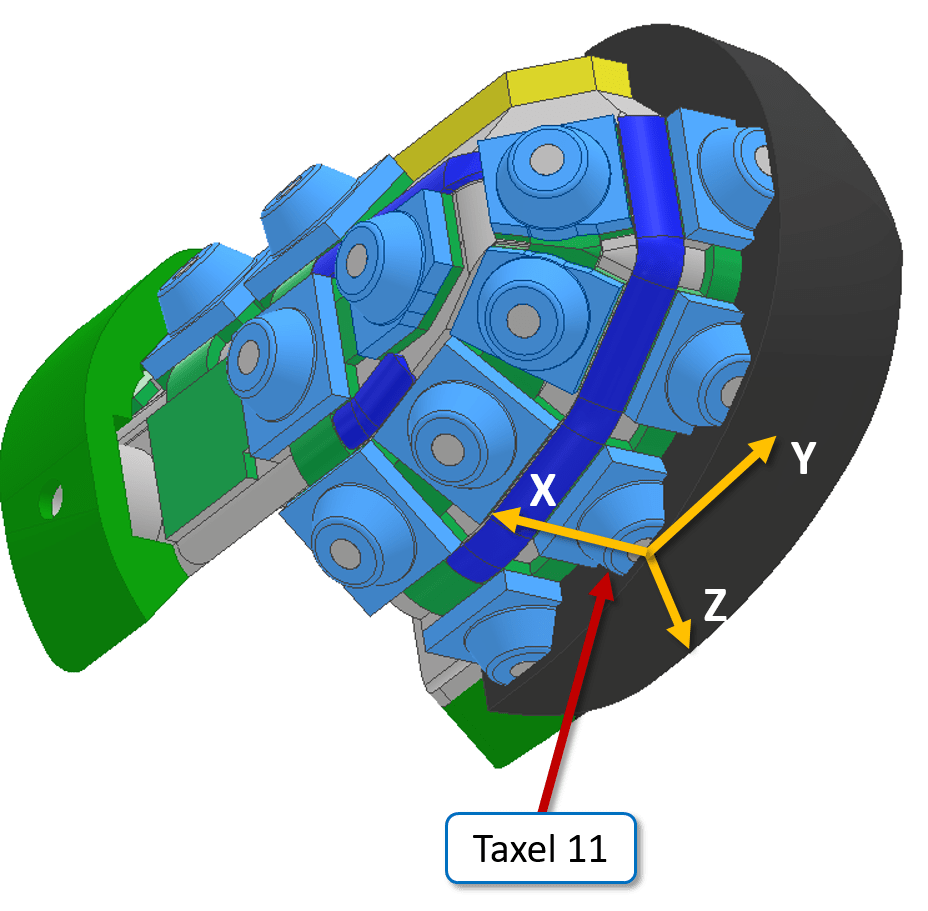}
      \caption{Taxel 11 used for calibration}
      \label{Taxel11}
   \end{figure}

The calibration methodology used to calibrate a single taxel will be illustrated in the upcoming section. As an example, taxel 11 is chosen for the calibration as it is often in contact with the flat surface mount during calibration. The location of the taxel can be seen in Fig.~\ref{Taxel11}. 
  
It was observed that by applying force randomly to calibrate the sensor, the sensor gave out results that cannot be used for the prediction of force measurement. Therefore, a simple yet systematic procedure was followed to calibrate the fingertip tactile sensor. 
\begin{figure*}[thpb]
     \centering
     \includegraphics[width=1.0\textwidth]{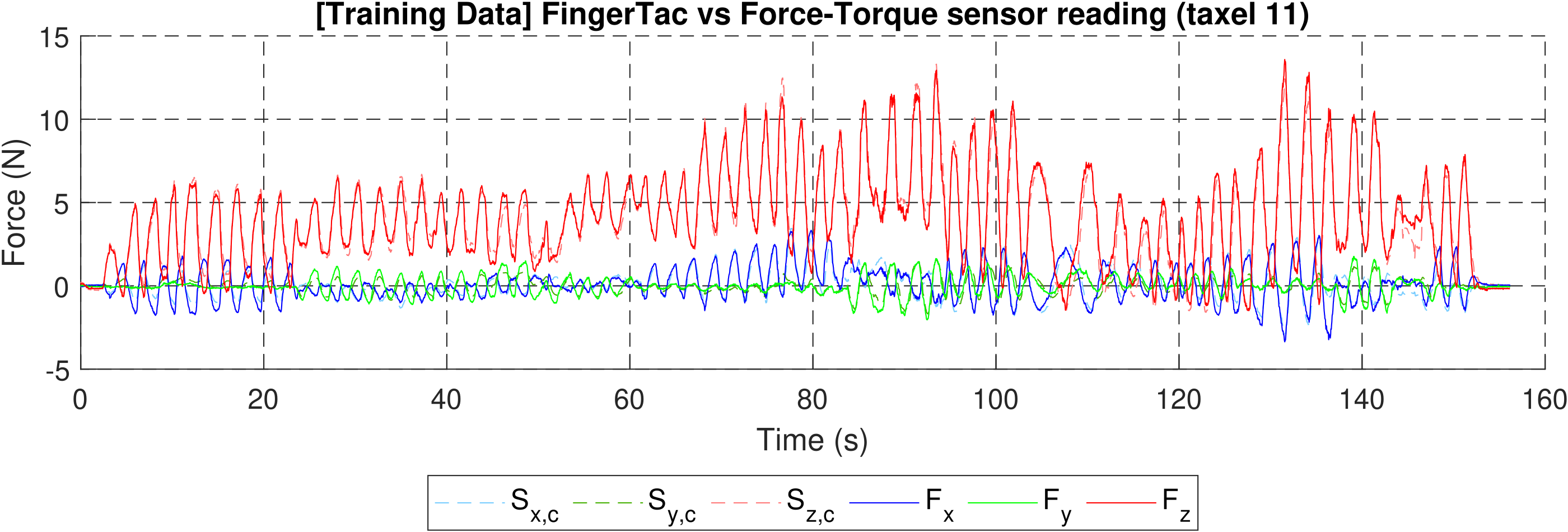}
     \caption{\textbf{Calibrated sensor readings and reference sensor readings during training}}
     \label{Training}
\end{figure*}

\begin{figure*}[thpb]
      \centering
      \includegraphics[width=1.0\textwidth]{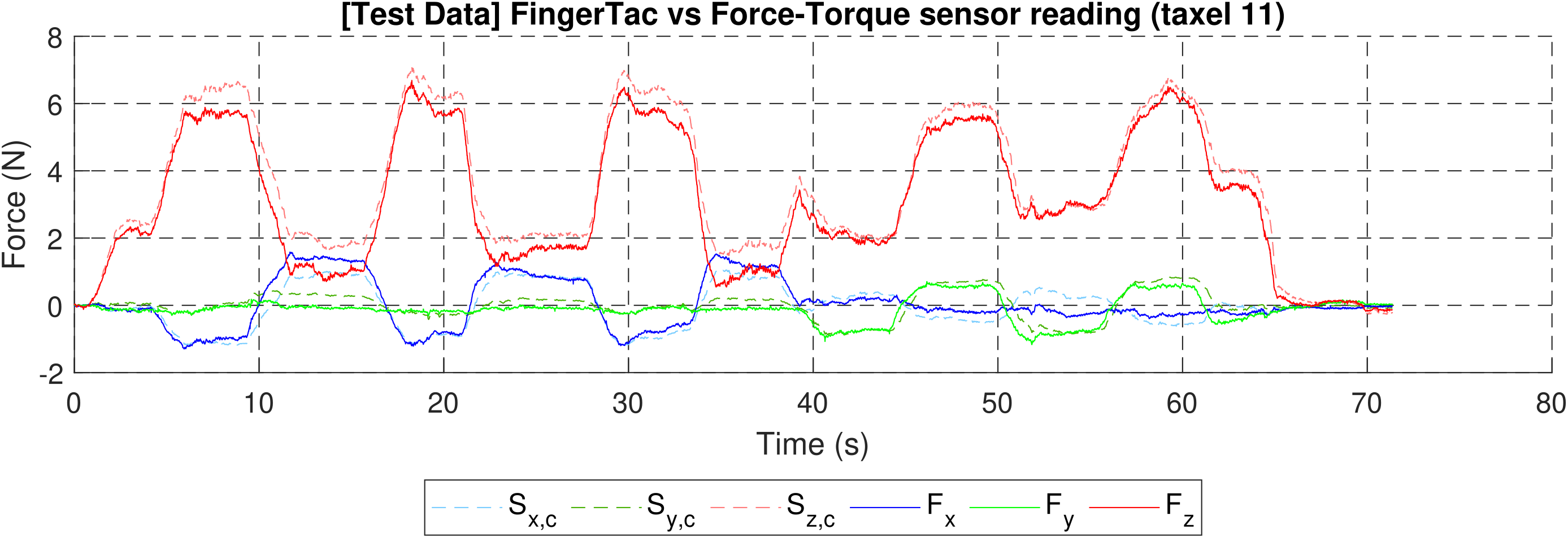}
      \caption{\textbf{Calibrated sensor readings and reference sensor readings during testing}}
      \label{Testing}
\end{figure*}
\subsubsection{Procedure to acquire training data}
At first, a test subject chosen for using the sensor for acquiring training data was asked to sit in a relaxed position with their elbows supported on the table. Then the subject was asked to wear the custom finger sleeve to prevent the sensor from slipping off the fingertip before putting on the fingertip tactile sensor. 

The subject was then asked to try out the sensor by applying normal force in the Z direction by pressing down on the flat surface mounted on top of the reference sensor and applying shear force by making slight movements in +/-X and +/-Y directions while maintaining the normal force down on the flat surface. 

A simple GUI was provided as feedback to view the resultant force values measured by the force-torque sensor. 


After trying out the sensor a couple of times the subject was asked to start the calibration process by 
applying normal and shear forces in an oscillatory fashion i.e. sequentially increasing and decreasing the magnitude of Z, -X, +X, -Y, and +Y forces. Fig.~\ref{Training} shows the training data captured by the subject while performing training data acquisition. For now we ask the reader to focus on the solid lines, which show the data from the force-torque sensor.

\subsubsection{Procedure to Acquire Test Data}
Similar to the training, in the data acquisition procedure the subject was asked to sit in a comfortable and relaxed position with the elbows supported on the table. The subject was then asked to apply sequentially increasing and decreasing magnitude of shear and normal forces in all axes. It was observed that it was difficult to apply forces of more than +/-2~N of shear force. 

Hence for the shear force, the subject was asked to apply forces between 0 and +/-2~N. While capturing the test data an additional instruction given to the subject was to maintain the magnitude of force which the subject can apply with ease between the above-mentioned magnitude limits for a period of around 5 seconds before changing the force magnitude. 

Fig.~\ref{Testing} shows the testing data captured by the subject while performing the test data acquisition procedure.

\section{Calibration Results}

The results of the calibration for the training data can be seen in Fig. \ref{Training} and test data can be seen in Fig. \ref{Testing}.
In both figures, the solid blue, green, and red lines represent the 3-axis force data measured by the force-torque sensor in X, Y, and Z directions, respectively. 
The dotted blue, green and red lines represent the predicted outputs of the regression model in the X, Y, and Z directions, respectively.
Looking at the result of calibration performed for training data in Fig.~\ref{Training}, we can see that the predictions from the polynomial regressions (dotted lines) are coincident with the actual force values (solid lines) for most time steps.
In Fig.~\ref{Testing} we can see the result for the test data. Here we can observe that the dotted lines tend to follow the actual force values measured by the force-torque sensor with some slight deviation. In the case of the Z-axis measurements, we can observe a slight overestimation. 
In X and Y axis the deviation from the actual measurements were lower compared to the Z-axis.  
The mean absolute error and root mean squared error per axis were calculated and are presented in Table~I. We can observe that both mean absolute error and root mean squared errors for all the axes are very low. The errors observed in the Z-axis is slightly higher than the error observed in the X and Y axis. This might be due to higher forces measured in the Z-axis. 
Overall, the low error values indicate that the tactile sensor can measure contact forces while handling objects. 



\begin{table}[h]
\caption{Mean Absolute Error and Root Mean Square Error per axis}
\centering
\resizebox{1.0\linewidth}{!}{
\begin{tabular}{|c||c|c|}
\hline
\textbf{Axis} & \textbf{Mean Absolute Error} & \textbf{Root Mean Squared Error}\\ \hline
X (N)         & 0.21                         & 0.28                         \\ \hline
Y (N)         & 0.16                         & 0.21                         \\ \hline
Z (N)         & 0.44                         & 0.52                         \\ \hline
\end{tabular}}
\end{table}

\section{Conclusion and Discussion}

In this paper we presented the design of a wearable fingertip tactile sensor named FingerTac. The design allows the user to measure tactile information not only while grasping an object but also while performing object manipulation.

A sensor taxel was calibrated to true force values and compared with an industrial-grade force-torque sensor. The calibration yielded good results proving that the sensor can be calibrated to measure 3-axis force applied by the human fingertips. However, we believe that for skill transfer the calibration of all taxels is not required, as FingerTac can be also applied to robot hands, i.e. the Allegro hand, and reproducing the uncalibrated sensor measurements might be sufficient. A lot of recent research in tactile sensing for robots uses uncalibrated sensor measurements \cite{Satoshi}.

In future work we plan to apply FingerTac to several human fingers and use it for skill transfer from humans to robots. Other possible future applications include skill transfer from human experts to non-experts and monitoring of humans, for example for quality control and optimization of products such as screw caps.




\bibliographystyle{IEEEtran}
\bibliography{Master_Reference}

\end{document}